# Generative Wind Power Curve Modeling Via Machine Vision: A Deep Convolutional Network Method with Data-Synthesis-Informed-Training

Luoxiao Yang, Long Wang, *Member, IEEE*, and Zijun Zhang, *Senior Member, IEEE*

*Abstract*—This study develops a novel data-synthesis-informed-training U-net (DITU-net) based method to automate the wind power curve (WPC) modeling without data pre-processing. The proposed data-synthesis-informed-training (DIT) process has following steps. First, different from traditional studies regarding the WPC modeling as a curve fitting problem, we renovate the WPC modeling formulation from a machine vision aspect. To develop sufficiently diversified training samples, we synthesize supervisory control and data acquisition (SCADA) WPC data based on a set of S-shape functions representing WPCs. These synthesized SCADA data and WPC functions are visualized as images, named the synthesized SCADA WPC and synthesized neat WPC, and paired as training samples. A deep generative model based on U-net is developed to approximate the projection recovering the synthesized neat WPC from the synthesized SCADA WPC. The developed U-net based model is applied into observed SCADA data and can successfully generate the neat WPC. Moreover, a pixel mapping and correction process is developed to derive a mathematical form depicting the neat WPC generated previously. The proposed DITU-net only needs to train once and does not require any data preprocessing in applications. Numerical experiments based on 76 WTs are conducted to validate the superiority of the proposed method via benchmarking against classical WPC modeling methods. To enable repeating presented research, we release our code and high resolution figures at https://github.com/IkeYang/DITU-net.

## ACRONYMS

| | |
|---|---|
| 4PLF | Four-parameter logistic function |
| 5PLF | Five-parameter logistic function |
| ADE | Adjusted double exponential |
| BSA | Backtracking search algorithm |
| CDF | Cumulative distribution function |
| CWS | Cut-in wind speed |
| DCAE | Deep convolutional autoencoder |
| DE | Double exponential |
| DIT | Data-synthesis-informed-training |
| DITDCAE | Data-synthesis-informed-training deep convolutional autoencoder |
| DITU-net | Data-synthesis-informed-training U-net |
| DKC | Domain knowledge correction |
| IDP | Insufficient data pattern |
| KNN | K nearest neighbor |
| MAE | Mean absolute error |
| MAPE | Mean absolute percentage error |
| ML | Machine learning |
| MS | Marker size |
| MSE | Mean square error |
| NP | Normal pattern |
| NRM | Newton-Raphson method |
| O&M | Operations and maintenance |
| RMSE | Root mean square error |
| RWS | Rated wind speed |
| SCADA | Supervisory control and data acquisition |
| SNN | Shallow neural network |
| SR | Spline regression |
| WMAPE | Weighted mean absolute percentage error |
| WPC | Wind power curve |
| WPCM | Wind power curve modeling |
| α-SS | α-shooting score |

## I. INTRODUCTION

To realize carbon neutrality by the mid-page of this century, wind energy will contribute more to the future energy portfolio. Meanwhile, as the wind industry has experienced a rapid growth over past decades, a significant proportion of wind power assets installed worldwide are aging, which poses an emerging demand of advanced technologies on the operations and maintenance (O&M) management side. Therefore, emerging technologies which benefit the wind farm O&M have become a new sweet spot in the wind energy domain. A wind power curve (WPC) directly pictures the wind speed and power output relationship and is useful to help address many issues in wind farm operations and power system operations, such as the wind power prediction [1, 2], load flow estimation [3], electricity market [4, 5] and wind energy assessment [6, 7]. Although WT manufacturers usually offer ideal WPCs of their WT models, it is widely known that real WPCs present quite different patterns due to numerous factors, such as the environmental changes, system degradations, control errors, etc.. Therefore, effective methods for modeling the real WPC via wind farm SCADA data are highly interesting and beneficial to both the wind industry and power grid.

In the literature, the wind power curve modeling (WPCM) based on wind turbine SCADA data has been vigorously discussed and various methods have been developed. Existing WPCM methods can be generally categorized into parametric models and non-parametric models [8]. The parametric model aims at approximating WPCs via explicit mathematical forms, such as the polynomial form, logistic function, and probabilistic

This work was supported in part by the National Natural Science Foundation of China Youth Scientist Fund Project with No. 52007160, in part by the Hong Kong Research Grants Council General Research Fund Projects with No. 11215418 and No. 11204419. (Corresponding author: Z. Zhang)

Luoxiao Yang and Zijun Zhang are with School of Data Science, City University of Hong Kong, Hong Kong SAR. (Email: zijzhang@cityu.edu.hk)
Long Wang is with Department of Computer Science and Technology, University of Science and Technology Beijing, Beijing, China.

model [9]. The linearized segmented model and the polynomial model were employed to fit the nonlinear part of WPC in [10] and [11] respectively. As several functions naturally have a similar S-shape curvature as the WPC, they have been applied to study the WPCM. In [12], the double exponential (DE) function and the adjusted double exponential function (ADE) were applied to perform WPCM. In [13] and [14], the four-parameter logistic function (4PLF) and the five-parameter logistic function (5PLF) were also applied into WPCM. In [15], a modified hyperbolic tangent model was introduced to characterize WPC. In [16], the author proposed a Gaussian cumulative distribution function (CDF) based model for studying WPCM. The probabilistic model studying WPCM treats all parameters of a model as random variables and employs the Bayesian inference to derive the corresponding posteriors. In [17], two statistical models, the heteroscedastic spline regression model and robust spline regression model were proposed for WPCM. To allow a higher flexibility of modeling various WPC shapes based on real SCADA data and capture the nonlinearity, non-parametric models, mainly ML based models, have been developed and applied. In [18], the spline regression (SR) method was introduced to conduct WPCM. In [19], the k nearest neighbor (KNN) was applied to accurately model the WPC in WT performance monitoring. In [20] and [21], the support vector regression and shallow neural network (SNN) were utilized to perform WPCM respectively. In [22], four data-driven models for WPCM were compared and results showed that the adaptive neuro-fuzzy-interference system model performed best.

Though existing WPCM methods have well proved the feasibility and effectiveness of modeling the true WPC from a data-driven perspective, two pain points still wait for further solutions. First, due to the quality of controllers, wind power curtailment, environmental changes, etc. [23], collected WT SCADA data might include noisy points which will significantly affect the performance of applied WPCM methods. Thus, existing data-driven methods for WPCM tasks usually require a detailed SCADA data preprocessing. Applied SCADA data preprocessing methods include the empirical approaches [17, 22], statistics-based models [24], and learning-based models [23, 25, 26]. As data pre-processing methods either require the strong domain expertise or involve the complicated tuning of hyperparameters, such practice is a time-consuming trial-and-error process and is usually ad-hoc. Moreover, most of the existing WPCM methods are dataset-specific, which means that the developed model is usually more effective for a specific WT during a specific period while applying such model into another WT or another time period might require a model renewal and even lead to a degraded performance. Thus, most of existing WPCM methods need to repeat the modeling process for different scenarios.

To develop a more general and adaptive paradigm for accurate WPCM free of any data pre-processing based on various WT datasets, in this study, we re-visit the WPCM problem from the image processing perspective, propose a new WPCM problem formulation considering image data derived from SCADA data, and develop an applicable image based generative WPCM method, a data-synthesis-informed-training U-net (DITU-net) based method. The proposed DITU-net based method includes two components, the data-synthesis-informed-training (DIT) strategy as well as the pixel mapping and correction. The DIT strategy is of two-fold, the real WPC image synthesis and WPC generative model development. To synthesize a real WPC image, we first regard a set of considered "S-shape" functions, $f_{wpc}$, as the WPC ground truths. Next, we randomly distribute data points around the curve of $f_{wpc}$ to synthesize the WPC image derived from SCADA data (SCADA WPC), $I_x$. The synthesized SCADA WPC images and the ground truth WPC are paired as training samples ($I_x, I_{wpc}$). A deep convolutional neural network, U-net [27], is then adapted to develop a U-net based model for extracting $I_{wpc}$ from $I_x$ based on synthesized training samples. In the implementation, the developed U-net based model is fed with the WPC images derived from real SCADA data and produce the corresponding $I_{wpc}$ with both a graphical visualization and a mathematical expression. To derive the mathematical form of the $I_{wpc}$, we first develop a process for mapping pixels to SCADA data points, and then employ a polynomial fitting with a domain knowledge based correction. To verify the effectiveness of the proposed DITU-net based WPCM method, four parametric models, DE, ADE, 4PLF and 5PLF, and two ML models, SNN and SR, are considered as benchmarks of the WPCM task. A comprehensive computational study has been conducted to demonstrate the effectiveness of the proposed DITU-net based WPCM method via using SCADA data collected from 76 commercial WTs under three application scenarios, WPCM with raw SCADA data, WPCM with roughly pre-processed SCADA data, and WPCM with carefully pre-processed SCADA data, as well as two patterns, normal pattern (NP) and insufficient data pattern (IDP). Computational results show that the proposed DITU-net based method, without any retaining, can achieve a comparable accuracy using raw SCADA data by comparing with the best benchmark using carefully pre-processed SCADA data and offer more accurate results than SNN with insufficient data.

Main contributions of this work are summarized as follows:
1) Novel problem formulation: A WPCM task formulated from the machine vision aspect is presented for the first time.
2) Methodological framework: A novel DITU-net based framework is developed to generate the neat WPC and WPC model expression based on SCADA WPC images.

3)  Model generalization: One DITU-net is trained and it shows a great potential of being generalized to different SCADA WPC data, which is verified based on a large set of WT SCADA data in this work.

4)  Model performance: The proposed method is free of the cumbersome SCADA data pre-processing required in traditional WPCM task. It directly uses raw SCADA data, performs as accurately as the best benchmark using carefully pre-processed SCADA data, and produces more accurate results when data are insufficient.

## II. Wind Power Curve Modeling Problem

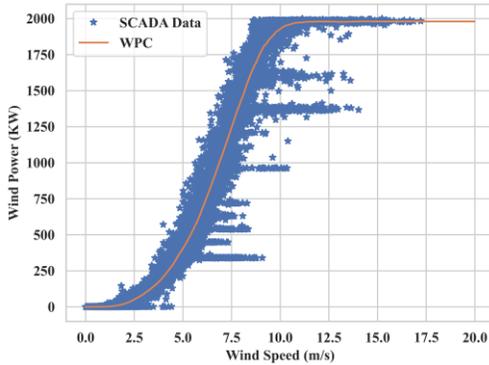

Fig. 1. A visualization of a neat WPC and the real WPC formed by SCADA data.

WPC is commonly regarded as a function describing the relationship between the wind speed and the generated wind power. Typically, the WPC adopts a piece-wise form as described in (1):

$$f_{wpc} = \begin{cases} 0, & x < x_{CutIn} \\ f(x), & x_{CutIn} \leq x < x_{Rated} \\ P_{Rated}, & x_{Rated} \leq x \end{cases} \quad (1)$$

where the $f(x)$ satisfies two boundary conditions as expressed in (2.1) and (2.2):

$$f(x_{CutIn}) = 0 \quad \frac{df(x)}{dx}\Big|_{x=x_{CutIn}} = 0 \quad (2.1)$$

$$f(x_{Rated}) = P_{Rated} \quad \frac{df(x)}{dx}\Big|_{x=x_{Rated}} = 0 \quad (2.2)$$

Fig. 1 visualizes a neat WPC in our expectation and the real WPC formed by WT SCADA data. It is observable that, due to various factors, such as the measurement errors, WT faults, wind curtailment, extreme weather conditions, etc., the raw WT data offers a rough curve formed by a swarm of data points. A common practice of modeling such a WPC via using SCADA data in the literature is to formulate the WPCM as a curve fitting/regression problem. Next, we will brief the traditional WPCM, discuss its limitations, and introduce the machine vision assisted WPCM formulation proposed in this work.

### A. Traditional wind power curve modeling

In the literature, the WPCM problem has been widely considered as a curve fitting problem (parametric model) or a regression problem (ML model). From this aspect, the observed SCADA data mainly serve as the input information for optimizing the parameters of a "S-shape" function or a ML model as described in (3):

$$p = \underset{p}{\operatorname{argmin}} \mathcal{L}(f_p(x), y) \quad (3)$$

where $f_p$ denotes a model used for depicting a WPC, $p$ denotes its parameters or weights, $\mathcal{L}$ denotes the loss function, which is usually a distance-based function (In the WPCM problem, $L_2$ norm is frequently chosen [19-21]), as well as $(x, y)$ denotes pairs of the collected wind speed and wind power output in SCADA data. Based on such problem formulation, vigorous discussions have been made on various model forms and machine learning techniques for obtaining WPCs with a higher quality [8, 9]. However, "noises" contained in the swarm of SCADA data can greatly affect the shape of curves obtained via fitting models into $(x, y)$ pairs, and thus a careful data preprocessing is usually required. In addition, the extracted WPC model $f_p$ is specific to the target WT only or maybe WTs sharing the same SCADA data distribution if exist. In other words, a WPC model developed via such traditional WPCM formulation is specific to data of the targeted WT and its form might no longer be effective to describe the WPC of another WT. To develop WPC models of a population of WTs using SCADA data via the traditional practice, it requires the turbine-wise repeated effort on data pre-processing and model fitting. It is interesting to revisit the WPCM problem and investigate the feasibility of developing a novel WPC modeling method which can be generalized to various WT SCADA data.

### B. A New Machine Vision Assisted Wind Power Curve Modeling Problem

A new formulation of the WPCM problem is proposed in this section, which first analyses raw WT SCADA data from the machine vision perspective and next returns a parametric form accurately reflecting the SCADA WPC. As the ground truth of a neat WPC representing the SCADA WPC is unavailable, we study our proposed WPCM method based on synthesized data to allow a rigorous evaluation and further verify the method based on real SCADA data. To synthesize a WPC visually like a SCADA WPC, we first identify a parametric form offering a "S-shape" curvature and randomly adding data points around such curve as described in (4):

$$x, y = f_\Lambda(f_{WPC}) \quad (4)$$

where the $f_{WPC}$ denotes the ground truth function picturing a neat WPC, $f_\Lambda$ denotes the described data synthesis function, and $(x, y)$ denotes pairs of the synthesized wind speed and wind power. We are able to synthesize a large volume of WPC images based on different pre-determined $f_{WPC}$ and the data synthesis process. Based on synthesized WPC images, our task is to obtain a data-driven model representing an inverse function of $f_\Lambda$, $f_\Lambda^{-1}$, which can derive the ground truth WPC

TABLE I.
DIFFERENCES AMONG THE TRADITIONAL WPCM, THE PROPOSED WPCM, AND THE WPP

| Task | WPCM | | WPP |
|---|---|---|---|
| | Machine vision assisted WPCM | Traditional WPCM | |
| Modeling objective | Obtain the wind power curve | | Obtain wind power predictions |
| Input | SCADA WPC (Image) | Wind speed and wind power pairs (Tabular-form data) | Historical sequence of SCADA attributes including the wind speed, wind power, rotor speed, wind direction, etc. (Tabular-form data) |
| Output | $f_{wpc}$ (A function) | | Predicted wind power (Numerical values) |
| Model | DITU-net | Classic curve fitting models/regression models | Regression models |
| Application | wind power prediction, electricity market and wind energy assessment, etc. | | Unit commitment [7], power market [8], power system scheduling [9], etc. |
| Connection | In several literature [10,11], the wind speed of the future is first predicted and then the future wind power is estimated via a WPC model. | | |

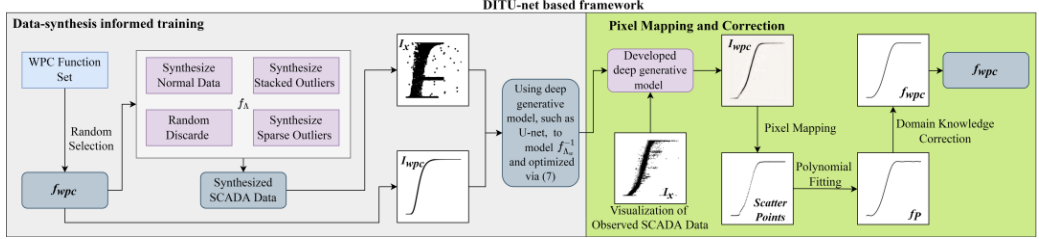

Fig. 2. The schematic diagram of the DITU-net based framework.

$$S_{WPC} = S_{DE} \cup S_{ADE}$$

$$S_{DE} = \left\{ f(\cdot) \middle| f(x) = e^{-t_1 e^{t_2 x}}, t_1 \sim U[10, 50], t_2 \sim U[-15, -8] \right\}$$

$$S_{ADE} = \left\{ f(\cdot) \middle| f(x) = e^{-e^{a_0 - a_1 x - a_2 x^2 - a_3 x^3}}, a_0 = 5, a_1 \sim U[a_2, b_2], a_2 \sim U[-15, 10], a_3 = 15 \right\} \quad (8)$$

function based on WPC images produced from the synthesized SCADA data as described in (5) as well as can be generalized to the observed real SCADA data:

$$f_{WPC} = f_{\Lambda}^{-1}(x, y) \quad (5)$$

Based on (5), the objective function of the considered data-driven modeling is then described in (6):

$$w = \underset{w}{\operatorname{argmin}} \mathcal{L}\left(f_{\Lambda_w}^{-1}(x, y), f_{WPC}\right) \quad (6)$$

where $f_{\Lambda_w}^{-1}$ denotes the inverse function with the weight of $w$ and $\mathcal{L}$ denotes the loss function. The (6) directly deals with values of $(x, y)$ while the local spatial relationship among $(x, y)$ pairs in a graphical visualization is not well incorporated. Let $I_x$ denote the image visualization of all pairs of $(x, y)$ and $I_{wpc}$ denote the visualization of $f_{WPC}$ respectively. We are able to extend the objective in (6) to a machine vision version as described in (7):

$$w = \underset{w}{\operatorname{argmin}} \mathcal{L}\left(f_{\Lambda_w}^{-1}(I_x), I_{wpc}\right) \quad (7)$$

We hereafter need to develop an effective method to address the modeling task stated in (7), learning a non-parametric form of $f_{\Lambda_w}^{-1}$ that can extract $I_{wpc}$ from $I_x$. Given a sufficiently large volume of training samples, we expect that the developed $f_{\Lambda_w}^{-1}$ can be applied into WPCM considering any SCADA data and can return high quality neat WPCs without any data pre-processing and model retraining. The (7) and (3) differ from two aspects. First, in (7), the objective function aims to learn an inverse projection from the SCADA WPC to WPC directly while, in (3), the objective function aims to learn a mapping from the wind speed to wind power. Secondly, in (7), the optimization function takes the synthesized SCADA WPC and the synthesized neat WPC as input pairs; however, in (3), the optimization function takes observed wind speed and wind power records as input pairs.

### C. Wind Power Curve Modeling Versus Wind Power Prediciton Modeling

Table I is developed to clearly differentiate three modeling problems, the proposed WPCM, traditional WPCM, and the wind power prediction modeling (WPPM). We can observe that the nature of WPCM and WPPM problems is different because they consider different inputs, different modeling purposes, and different modeling outcomes. The traditional and our study investigates two completely different WPCM paradigms: the traditional studies consider WPCM from fitting models into SCADA data while our study proposed WPCM from the WPC image generation and analysis direction.

### III. THE DATA-SYNTHESIS-INFORMED-TRAINING U-NET BASED WIND POWER CURVE MODELING METHOD

To address the proposed WPCM formulation, we develop a DITU-net based methodological framework and describe its principle next. Please note that the U-net is chosen as a base model in this study because it is a classical convolutional autoencoder, which offers a symmetric architecture for inverse engineering and has widely demonstrated its successes in medical imaging analytics, especially in feature engineering. Theoretically, any generative deep network offering similar or better performance than the studied U-net on learning the inverse form of a convolutional process can fit into the proposed framework. Fig. 2 provides a schematic diagram of the proposed DITU-net based framework, which contains two phases, the DIT phase and the pixel mapping and correction phase. In the DIT phase, based on the setup of the proposed WPC formulation in Section II, a WPC function generator randomly selects a WPC function $f_{WPC}$ from a pre-determined function set. Considering various types of SCADA data swarms displayed in SCADA WPC as mentioned in [25], a $f_\Lambda$ is designed and output the synthesized SCADA data points based on a selected $f_{WPC}$. The synthesized SCADA data and selected $f_{WPC}$ are then visualized as image pairs and denoted as $I_x$ and $I_{wpc}$ respectively. Due to the powerful feature engineering capability of the U-net, it is applied to model $f_{\Lambda_w}^{-1}$ and optimized via considering (7) as the loss function. In the pixel mapping and correction phase, the observed SCADA data are first applied to produce an image of the SCADA WPC, $I_x$, which is then fed into the developed U-net to generate a neat WPC, $I_{wpc}$. Based on the mapping relationship between the pixel and the real-world SCADA data points as well as the domain knowledge, the $I_{wpc}$ is finally converted into a parametric representation, $f_{wpc}$.

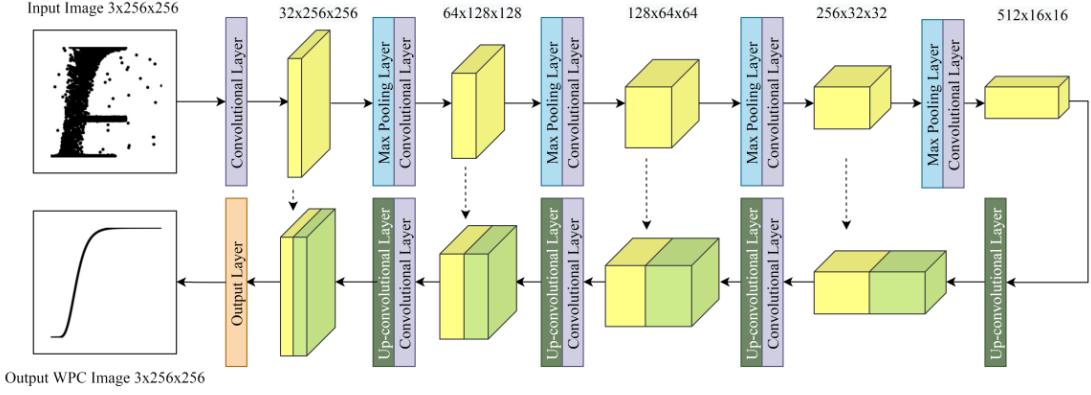

Fig. 3 The architecture of the U-net backbone utilized in data-synthesis-informed-training

We next sequentially introduce details in the DIT strategy, in the pixel mapping and correction, as well as in the training and testing scheme of the proposed DITU-net.

### A. Data-synthesis-informed-training strategy

To develop a data-driven model approximating $f_{\Lambda_w}^{-1}$ via (7), training pairs, $I_x$ and $I_{wpc}$, are needed. Yet, we need to address two challenges relating to data. On one hand, it is difficult to obtain a sufficient diversity of pairs of $I_x$ and $I_{wpc}$ via real SCADA dataset. On the other hand, the ground truth of the neat WPC $I_{wpc}$ based on a SCADA WPC is unavailable. Thus, in this research, we develop a data synthesis process, $f_\Lambda$, to generate sufficient pairs of training samples via simulating the SCADA WPC and then utilize synthesized data to develop a U-net based model serving as the $f_{\Lambda_w}^{-1}$.

#### 1) Data synthesis

A data synthesis process, $f_\Lambda$, is developed to simulate a SCADA WPC via randomly adding swarms of data points around a "S-shape" function representing a neat WPC. Based on the developed DIT data synthesis process, pairs of WPC images $I_{wpc}$ and the simulated SCADA data images $I_x$ are generated. For simplicity, in this research, we choose the double exponential (DE) function and the adjusted double exponential (ADE) function [12] as illustrative examples and assume that they are two functions describing ground truths of neat WPCs.

Thus, the DE and ADE form a WPC function set $S_{WPC}$ as expressed in (8), where $x$ denotes the normalized wind speed and $U$ denotes a uniform distribution. In (8), the upper and lower bound settings of each parameter are given based on our preliminary trials, which achieve a tradeoff between the stability and diversity of generated $f_{WPC}$.

In [25], the observed SCADA data are categorized into three patterns, the normal data, the stacked outliers, and the sparse outliers. Based on the generated $f_{WPC}$, three patterns of SCADA data are sequentially simulated via (9):

$$Normal\ Data = (x, \epsilon * \phi\left(derivative\left(f_{WPC}(x)\right)\right) + f_{WPC}(x))$$

$$Stacked\ Outliers = (x, \epsilon * std + f_{WPC}(x))$$

$$Sparse\ Outliers = (a, b)$$

$$derivative = \frac{df_{WPC}(x)}{dx}$$

$$\epsilon \sim N(0,1), a, b \sim U[0,1] \tag{9}$$

where $\phi$ denotes the variance projection (VP) function. Here, we assume that the variance of $Normal\ Data$ is proportional to the first-order derivative of the $f_{WPC}(x)$. However, directly utilizing the derivative of $f_{WPC}(x)$ could result in a situation that normal data are unavailable. Thus, an empirical design of the $\phi$ is proposed in (10) to avoid this situation:

$$\phi(d) = \begin{cases} \eta(d), & Normaliz(d) < 0.7 \\ (1 - (\eta(d) - 1)^4)^{0.5}, & Normaliz(d) \geq 0.7 \end{cases} \tag{10}$$

where $\eta(\cdot)$ denotes a minmax normalization. The finally synthesized SCADA WPC is a concatenation of previously mentioned three patterns of data. To simulate scenarios of

---

**Algorithm 1:** Data synthesis process

**1: Input**: The number of total synthesized samples $n$, the number of three types SCADA data, $n_{normal}$, $n_{stacked}$, $n_{sparse}$.

2:   **for** $i$=1 to $n$

3:      Generate $f_{WPC}(\cdot)$ from $S_{WPC}$.

4:      $SSD = [\ ]$

5:      Synthesize $n_{normal}$ normal data $d_{norm}$ based on (9).

6:      $SSD = [SSD, d_{norm}]$

7:      Synthesize $n_{stacked}$ stacked outliers $d_{stacked}$ based on (9).

8:      $SSD = [SSD, d_{stacked}]$

9:      Synthesize $n_{sparse}$ sparse outliers $d_{sparse}$ based on (9).

10:     $SSD = [SSD, d_{sparse}]$

11:     Randomly discard $SSD$.

12:     Display $f_{WPC}(\cdot)$ as $I_{wpc}$ and $SSD$ as $I_x$. Save $(I_x, I_{wpc})$.

**13:   end for**

**14: Output**: A dataset with $n$ training pairs.

---

**Algorithm 2:** Data-synthesis-informed-training

**1: Input**: The hyperparameter controlling data synthesis ($n$, $n_{normal}$, $n_{stacked}$, $n_{sparse}$), the number of U-net training iteration $n_{iter}$

2:   Synthesize $n$ training pairs ($I_x$, $I_{wpc}$) via **Algorithm 1**.

3:   **for** epoch $i$=1 to $n_{iter}$

4:      $loss_{Unet} = \left\| I_{wpc} - Unet(I_x) \right\|_2^2$

5:      Update U-net via a gradient based optimizer

**6:   end for**

**7: Output**: Developed U-net

---

WPCs containing insufficient data, the synthesized SCADA data points exceeding a randomly set wind power or wind speed level will be discarded. Finally, the synthesized SCADA data points and the $f_{WPC}$ are displayed graphically and saved as images for model training. The Pseudo code of the proposed data synthesis process in DIT is offered in **Algorithm 1**, where $SSD$ is short for Synthesized SCADA Data. In Appendix I, four randomly selected synthesized training pairs are visualized.

### 2) A U-net Based Model Development

A U-net is a DNN adopting the autoencoder architecture and is firstly proposed for the cell detection and segmentation in [27]. Fig. 3 presents the architecture of the U-net utilized in this paper, where the number on the top specifies the configuration of feature maps and the dash line denotes a copy and concatenate operation (also called skip-connection). It is observable that the considered U-net is composed of four basic layers, the convolutional layer, the max pooling layer, the up-convolutional layer, and the output layer. Each convolutional layer repeats three successive operations, the $3\times3$ convolutions, batch normalization [28] and ReLU activation, twice. The max pooling layer utilizes a $2\times2$ pooling operation for down sampling and the up-convolutional layer utilizes a $2\times2$ transpose convolutional operation for up sampling. The final output layer employs a single convolutional operation aiming to map the feature map to the RGB image space. The encoder part of the U-net conducts four times of down sampling operations to reduce the feature size from $256\times256$ to $16\times16$. Accordingly, the decoder part conducts four times of up sampling operations to restore the original image resolution from low-dimensional high-level semantic features output by the encoder. In addition, the skip connection, which is the main advantage of U-net, is utilized between the same stage of the encoder and decoder to ensure that the final feature map could fuse both low-level and high-level features. To extract the neat WPC from the SCADA WPC, both high-level features (the tendency of SCADA WPC) and low-level features (local noise) are important. Thus, the U-net is adopted in this paper.

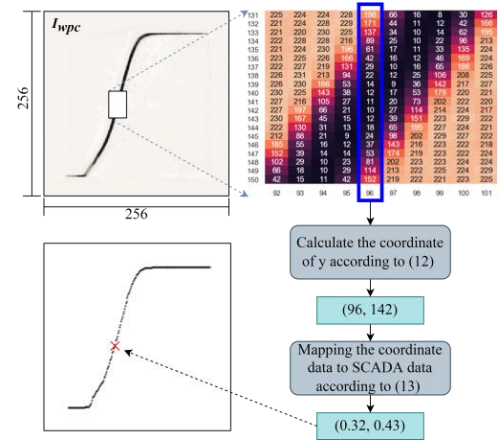

Fig. 4. The pixel mapping process.

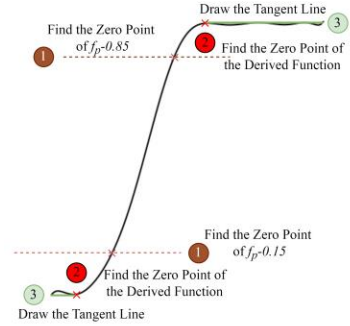

Fig.5 The domain knowledge correction process.

The mean square error (MSE) loss described in (11) is considered as the $\mathcal{L}$ in (7):

$$\mathcal{L} = \left\| I_{wpc} - Unet(I_x) \right\|_2^2 \tag{11}$$

The pseudo code of the whole DIT process is offered in **Algorithm 2**.

### B. Pixel mapping and correction

After the DIT, the developed U-net is capable to perform online WPCM based on real SCADA data. The observed SCADA data of the wind speed and wind power output are first applied to produce images $I_x$, a SCADA WPC, and then sent to the developed U-net to obtain the output $I_{wpc}$, a neat WPC. To derive $f_{wpc}$ from $I_{wpc}$, in this section, we present a pixel mapping and correction process, which contains three steps, the pixel mapping, polynomial fitting, and domain knowledge correction (DKC). Before elaborating the proposed pixel mapping and correction process, we define the coordinate of the pixel in an image as the pixel coordinate, which is composed of two integers representing the relative position of a pixel in an image, and we define the coordinate of normalized SCADA data as the SCADA coordinate, which is composed of two floats ranging from 0 to 1 representing the normalized wind speed and the corresponding normalized wind power respectively.

Fig. 4 illustrates the pixel mapping process. It is observable that the output $I_{wpc}$ is first converted into a grey scale image with a size of $256 \times 256$. The feasible pixel coordinate of $x$ is ranged from 32 to 229 and $y$ is ranged from 31 to 227. We note $227 - 31 + 1 = 197$ as the validated resolution $R_s$ of $I_{wpc}$. The inconsistency of the range between $x$ and $y$ is caused by the plot package utilized in this paper. Thus, there are $229 - 32 + 1 = 198$ feasible coordinate pairs to be decided. In this research, the coordinate of $y$ is decided based on the minimum value of the pixel in one column as expressed in (12):

$$y = \text{mean}\left(\underset{i}{\text{argmin}}\, I_{wpc}[x, i]\right) \quad (12)$$

To map the pixel coordinate into SCADA data coordinate, the coordinate pairs are converted via (13):

$$x = \frac{x - 32}{229 - 32}$$
$$y = 1 - \frac{y - 31}{227 - 31} \quad (13)$$

After pixel mapping, there are 198 normalized SCADA data points, which are utilized to fit an n-order polynomial as described in (14):

$$\min \sum_{i=1}^{198} (f_p(x_i) - y_i)^2 + \lambda \|\boldsymbol{a}\|_2^2$$
$$s.t. \quad f_p(x) = \sum_{i=0}^{n} a_i x^i \quad (14)$$

where the $\lambda$ is the penalty term.

However, due to the high flexibility of the polynomial in fitting a curve and the error in the pixel mapping process, the fitted $f_p$ could conflict with the domain knowledge in terms of the cut-in and rated wind speed as shown in Fig. 5. Thus, in this research, we propose a DKC process to ensure that the output $f_{wpc}$ meets WPC definition described in (1), (2.1) and (2.2). The main purpose of the DKC is to determine the cut-in wind speed (CWS), $x_{CutIn}$, and the rated wind speed (RWS), $x_{Rated}$, in the fitted curve, which makes the first-order derivative of $f_p$ be equal to zero at CWS and RWS. Considering that it is convenient to drive the explicit expression of a polynomial derivative, in this paper, the Newton-Raphson method (NRM) is utilized to decide CWS and RWS. Fig. 5 presents the details of the DKC, which can be summarized into three steps. First, a bisection is utilized to decide an appropriate initial point of the NRM for calculating CWS and RWS respectively as described in (15):

$$x_{init\_c} = Bisection(f_p - c) \quad (15)$$

where $c$ is set to 0.85 for RWS and is set to 0.15 for CWS.

Next, based on $x_{init}$, the corresponding CWS $x_{CutIn}$ and RWS $x_{Rated}$ are calculated via (16):

$$x_{CutIn \setminus Rated} = NRM(f_p', x_{init\_0.85 \setminus 0.15}) \quad (16)$$

where $f_p'$ is the first-order derivative of $f_p$.

Finally, a tangent line is drawn to meet the first-order continuity at the point of tangency. The output $f_{wpc}$ is then expressed in (17):

$$f_{wpc} = \begin{cases} f_p(x_{CutIn}), & x < x_{CutIn} \\ f_p(x), & x_{CutIn} \le x < x_{Rated} \\ f_p(x_{Rated}), & x_{Rated} \le x \end{cases} \quad (17)$$

We next explore from a theoretical aspect that whether the WPC model expression produced after the pixel mapping and correction process can meet the WPC definition constraints described in (2.1) and (2.2).

***Assumption 1.*** We assume that the order of the $f_p(x)$ is sufficient to model the curvature of SCADA data displayed in a two-dimensional normalized wind speed and wind power scatter diagram, and (18) holds.

$$0 \le f_p(x) \le 1 \qquad f_p(x_{CutIn}) \le \frac{l-1}{R_s} \quad (18)$$

where $l$ denotes the width of the plotted line and $R_s \in N^*$ denotes the validated image resolution.

***Assumption 2.*** We assume that turbines are healthy and normally controlled so that the most frequent value of wind power with wind speed records higher than RWS should be equal to the rated wind power $P_{Rated}$.

***Assumption 3.*** Assume the pixel value of the curved part of $I_{wpc}$ is inversely related to the density of the $I_x$ and, based on ***Assumptions 1*** and ***2***, (19) holds.

$$P_{Rated} - \frac{l-1}{R_s} \le f_p(x_{Rated}) \le P_{Rated} + \frac{l-1}{R_s} \quad (19)$$

TABLE II
DESCRIPTION OF DATASETS UTILIZED IN THIS STUDY

| Dataset | Location | Number of WTs | Collected Period | No. of Data Points of each WT |
|---|---|---|---|---|
| QH1 | Qinghai | 8 | Mar. 5th, 2016 - May 5th, 2016 | 153388 |
| QH2 | Qinghai | 8 | Nov. 5th, 2016 -Jan. 5th, 2017 | 87840 |
| SX | Shaanxi | 16 | Jan. 1th, 2015 -Aug. 12th, 2014 | 31267 |
| NX | Ningxia | 30 | Feb. 1th, 2014 - Jun. 27th, 2014 | 20433 |
| LN | Liaoning | 10 | Apr. 1th, 2015 - Jun. 10, 2015 | 9895 |
| HB | Hebei | 4 | Apr. 1th, 2015 - Jun. 6, 2015 | 9499 |

*The sampling interval of QH1 and QH2 is 1 min and that of other datasets is 10 min

---

**Algorithm 3:** The training and testing scheme of the proposed DITU-net

1: **Training**
2:　Based on Algorithm 2 train U-net.
3: **Testing**
4:　Visualize the observed SCADA data noted as $I_x$.
5:　Output $I_{wpc} = Unet\ (I_x)$.
6:　Based on (12-13), calculate scatter points $(x, y)$.
7:　Fit a polynomial $f_p$ based on (14).
8:　Output $f_{wpc}$ based on (15-17).

---

**Proposition 1**. Based on assumptions 1 and 3, it is guaranteed that the $f_{wpc}$ defined in (17) can meet two boundary conditions in (2.1) and (2.2) if $R_s$ approaches infinity.

*Proof:* Considering the WPC curve defined in (1), the value of the $f_p'$ is equal to zero at $x_{CutIn\setminus Rated}$. Thus, we have (20):

$$\frac{df_{wpc}(x)}{dx}\Big|_{x=x_{CutIn}} = f_p'(x_{CutIn}) = 0$$
$$\frac{df_{wpc}(x)}{dx}\Big|_{x=x_{Rated}} = f_p'(x_{Rated}) = 0 \quad (20)$$

By considering **Assumption 1**, based on (17) and (18), we can obtain (21):

$$0 \le f_{wpc}(x)|_{x \le x_{CutIn}} = f_p(x_{CutIn}) \le \frac{l-1}{R_s} \quad (21)$$

If $R_s$ in (21) approaches infinity, then (22) holds:

$$\lim_{R_s \to +\infty} f_{wpc}(x)|_{x \le x_{CutIn}} = 0 \quad (22)$$

The produced WPC model in (17) meets the boundary condition (2.1).

By considering **Assumption 3**, if $R_s$ in (19) approaches infinity, then (23) holds:

$$\lim_{R_s \to +\infty} f_{wpc}(x)|_{x \ge x_{Rated}} = P_{Rated} \quad (23)$$

The produced WPC model in (17) meets the boundary condition (2.2).

This establishes **Proposition 1**.

The pseudo code of the overall training and testing scheme of the proposed DITU-net based WPCM is offered in **Algorithm 3.**

## IV. COMPUTATIONAL EXPERIMENT

### A. Dataset

In this research, six SCADA datasets collected from commercial wind farms in Mainland China are utilized to study the WPCM problem. The usage of collected SCADA data is of two-fold. First, we analyze patterns of collected SCADA data to achieve a high-quality synthesis of real WPCs. In addition, we verify the proposed method and considered benchmarks on the WPCM task using collected SCADA data. Details of the considered datasets are reported in Table II.

### B. Benchmarking Models and Model Parameter Setup

In this research, six classical benchmarks, DE, ADE, 4PLF, 5PLF, SNN, SR, and DIT deep convolutional autoencoder (DITDCAE), are considered to verify the advantage of the proposed DITU-net. The deep convolutional autoencoder (DCAE) adopts the same convolutional layer settings as the U-net except the copy and concatenate operation. The backtracking search algorithm (BSA) [29] is employed to optimize parameters of the parametric model including DE, ADE, 4PLF, and 5PLF as recommended in [17]. The number of hidden neurons of SNN is set to 50 based on a preliminary trial.

In the DIT process, the number of total synthesized SCADA data $n$ is set to 4000 with $n_{normal}$, $n_{stacked}$, and $n_{sparse}$ set to 1000, 150, and 250, respectively. To eliminate the inconsistency between the total number of data points in synthesized SCADA data and that of observed SCADA data, the marker size (MS) of $I_x$ of the observed SCADA data should be proportional to the MS of $I_x$ of the synthesized SCADA data as described in (24)

$$K * MS_{DIT}^2 * 1400 = MS_{Test}^2 * n_{data} \quad (24)$$

where $n_{data}$ is the number of data points in the observed SCADA data, $MS_{DIT}^2 = 6$, as well as $K$ is the proportionality coefficient and is set to 0.07 based on the grid search. Then, the detailed visualization settings are reported in Table III.

TABLE III
VISUALIZATION SETTINGS

| Setting | DIT | Testing |
|---|---|---|
| Figure size | 256*256*3 | 256*256*3 |
| Marker Size | 6 | $\sqrt{K * 1400 * 6^2/n_{data}}$ |
| Color | Black | Black |

### C. Three application scenarios and two patterns

**Definition 1. The training set.**
In this study, we consider three application scenarios and two patterns to verify the proposed method and benchmarks. Please note that, once the DIT process is complete, the DIT-based method is applied into data without any further training. We just visualize the training set, feed it into the developed DIT-based method, and obtain the $f_{wpc}$.

**Scenario I. WPCM with raw SCADA Data:** Scenario I (S1) is analogous to the online WPCM task, which directly considers observed SCADA data as the model training source without any pre-processing.

**Scenario II. WPCM with roughly pre-processed SCADA data:** Scenario II (S2) describes WPCM tasks considering a low latency so that SCADA data can be roughly pre-processed and then considered in the training.

**Scenario III. WPCM with carefully pre-processed SCADA data:** Scenario III (S3) represents the offline WPCM task, which allows to consider SCADA data carefully pre-processed by domain experts in model training.
**The normal pattern (NP):** The model could access all training data.

**The insufficient data pattern (IDP):** The model could only access partial training data. It could happen in the wind farm start-up phase, which has insufficient historical records.

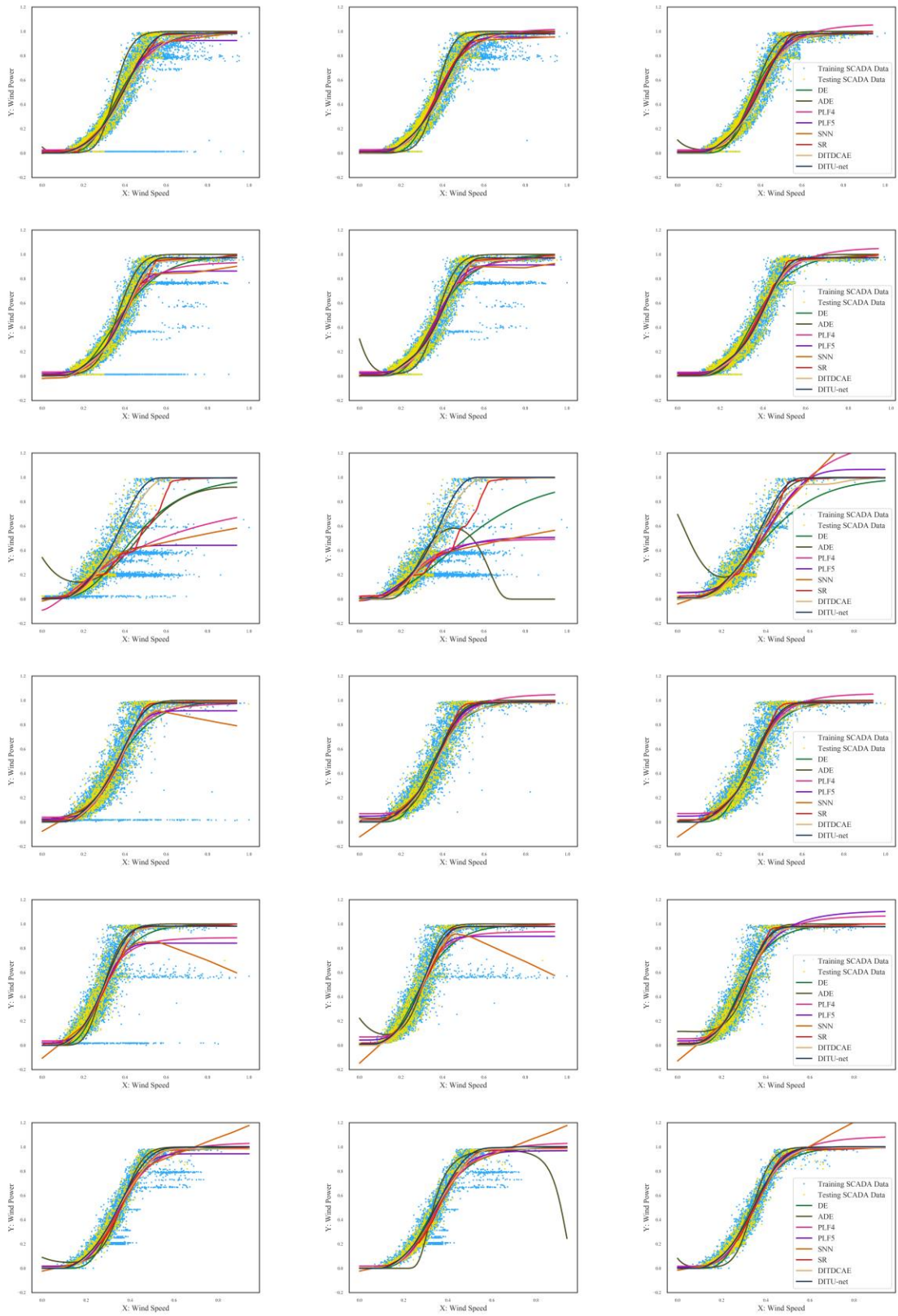

(a) S1          (b) S2          (c) S3

Fig. 6. Illustrations of WPCM results.

TABLE IV
THE RESULTS OF THE TESTING SET

| Method | DE | ADE | PLF4 | PLF5 | SNN | SR | DITDCAE | DITDCAE w/o DKC | DITU-net | DITU-net w/o DKC |
|---|---|---|---|---|---|---|---|---|---|---|
| RMSE(S1) | 0.08423 | 0.09636 | 0.07975 | 0.07959 | 0.07773 | 0.07610 | 0.07659 | 0.07659 | **0.06960** | 0.07062 |
| RMSE(S2) | 0.08194 | 0.09000 | 0.07567 | 0.07640 | 0.07355 | 0.07319 | 0.07632 | 0.07631 | **0.06960** | 0.07066 |
| RMSE(S3) | 0.07946 | 0.09138 | 0.07102 | 0.07213 | **0.06838** | 0.06861 | 0.07336 | 0.07335 | 0.06893 | 0.06889 |
| MAE(S1) | 0.06132 | 0.07258 | 0.05728 | 0.05689 | 0.05492 | 0.05037 | 0.05254 | 0.05246 | **0.04688** | 0.04710 |
| MAE(S2) | 0.06024 | 0.06605 | 0.05421 | 0.05419 | 0.05102 | 0.04718 | 0.05232 | 0.05223 | **0.04684** | 0.04709 |
| MAE(S3) | 0.05893 | 0.06755 | 0.05087 | 0.05049 | 0.04626 | **0.04399** | 0.05049 | 0.05040 | 0.04616 | 0.04602 |
| MAPE(S1) (%) | 15.06 | 18.57 | 14.66 | 14.57 | 14.28 | 14.07 | 13.99 | 13.99 | **12.80** | 12.87 |
| MAPE(S2) (%) | 14.99 | 16.91 | 13.95 | 13.71 | 13.27 | 13.22 | 13.95 | 13.94 | **12.82** | 12.88 |
| MAPE(S3) (%) | 14.59 | 17.31 | 13.41 | 13.02 | 12.5 | **12.45** | 13.59 | 13.58 | 12.68 | 12.67 |
| WMAPE (S1) (%) | 17.46 | 20.79 | 16.21 | 16.03 | 15.49 | 14.17 | 15.03 | 15.00 | **13.36** | 13.43 |
| WMAPE (S2) (%) | 17.15 | 18.98 | 15.38 | 15.32 | 14.41 | 13.29 | 14.97 | 14.94 | **13.35** | 13.42 |
| WMAPE (S3) (%) | 16.84 | 19.43 | 14.53 | 14.30 | 13.15 | **12.50** | 14.44 | 14.41 | 13.14 | 13.10 |
| 0.05-SS (S1) (%) | 55.24 | 47.25 | 61.06 | 60.24 | 61.63 | 65.69 | 63.87 | 63.87 | **67.91** | 67.87 |
| 0.05-SS (S2) (%) | 55.40 | 52.77 | 63.30 | 65.42 | 68.30 | 63.96 | 63.96 | **67.85** | 67.84 |
| 0.05-SS (S3) (%) | 55.27 | 51.59 | 65.18 | 65.36 | 68.59 | **70.06** | 64.93 | 64.93 | 68.34 | 68.35 |
| 0.10-SS (S1) (%) | 79.61 | 73.50 | 81.28 | 81.64 | 82.54 | 83.99 | 83.33 | 83.33 | **86.26** | 86.22 |
| 0.10-SS (S2) (%) | 80.59 | 77.32 | 83.80 | 83.43 | 84.82 | 85.09 | 83.45 | 83.45 | **86.26** | 86.21 |
| 0.10-SS (S3) (%) | 81.87 | 76.43 | 85.86 | 85.01 | 86.73 | **86.85** | 84.55 | 84.55 | 86.50 | 86.53 |
| 0.15-SS (S1) (%) | 90.70 | 87.54 | 91.44 | 91.63 | 92.25 | 92.20 | 92.50 | 92.50 | **93.91** | 93.87 |
| 0.15-SS (S2) (%) | 91.54 | 89.48 | 92.80 | 92.54 | 93.01 | 92.70 | 92.56 | 92.56 | **93.90** | 93.87 |
| 0.15-SS (S3) (%) | 92.59 | 89.25 | 93.99 | 93.49 | **94.16** | 93.93 | 93.33 | 93.33 | 94.09 | 94.09 |

TABLE V
THE AVERAGE TIME CONSUMPTION FOR WPCM

| Method | DITU-net | DITU-net w/o DKC | DITDCAE | DITDCAE w/o DKC | DE | ADE | PLF4 | PLF5 | SNN | SR |
|---|---|---|---|---|---|---|---|---|---|---|
| Running Time (s) | 0.01342 | 0.003263 | 0.01354 | 0.003345 | 0.4413 | 1.737 | 2526 | 2529 | 0.9070 | 0.002768 |

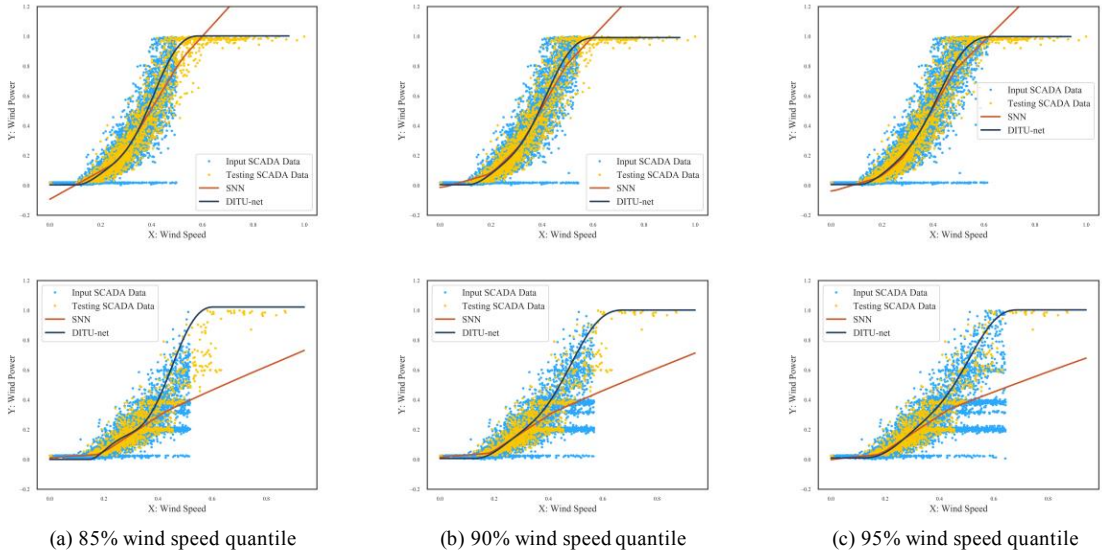

(a) 85% wind speed quantile      (b) 90% wind speed quantile      (c) 95% wind speed quantile

Fig. 7. The WPCM results of the proposed DITU-net with insufficient data pattern

**Definition 2. The testing set:** 20% of data points from the data utilized in S3 are randomly filtered out and selected as the testing dataset.

### D. Model Assessment Metrics

Let $x_i$ and $y_i$ denote the $i^{th}$ ground truth wind speed and wind power in the testing dataset respectively. Following metrics, the

Root Mean Square Error (RMSE), Mean Absolute Error (MAE), Mean Absolute Percentage Error (MAPE), Weighted Mean Absolute Percentage Error (WMAPE), and $\alpha$-shooting score ($\alpha$-SS) described in Eqs. (25) − (29) are utilized in this research to evaluate the performance of WPCM methods. To exclude the poor performance of MAPE on targets of values close to zero, only SCADA data with wind speed records higher than the CWS are considered in computing the MAPE.

$$RMSE = \sqrt{\frac{1}{n}\sum_{i=1}^{n}(f_{WPC}(x_i) - y_i)^2} \qquad (25)$$

$$MAE = \frac{1}{n}\sum_{i=1}^{n}|f_{WPC}(x_i) - y_i| \qquad (26)$$

$$MAPE = \frac{100\%}{n}\sum_{i=1}^{n}\left|\frac{f_{WPC}(x_i) - y_i}{y_i}\right| \qquad (27)$$

$$WMAPE = 100\% \times \frac{\sum_{i=1}^{n}|f_{WPC}(x_i) - y_i|}{\sum_{i=1}^{n}|y_i|} \qquad (28)$$

$$\alpha - SS = 100\% \times \frac{\mathbb{1}(|f_{WPC}(x_i) - y_i| \leq \alpha)}{n} \qquad (29)$$

where $\mathbb{1}$ is the indicator function.

*E. Computational Results*

Fig. 6 visualizes the WPCM results of six randomly selected WTs under three application scenarios in NP (we only plot the legend at S3 to make figures clearer). It is observable that the proposed DITU-net can well model the WPC regardless of the observed SCADA data quantity. The other benchmarks could offer suitable WPC modeling results under S3; however, if the observed SCADA data points incorporate noisy points or the application scenario requires the low latency, they failed to model WPCs accurately.

Table IV reports the testing results of considered methods under different application scenarios in NP. Considering that the testing set is sampled from data utilizes in S3, the accuracy of the proposed DITU-net could not exceed ML based models, which are directly trained with data utilized in S3. Even though, the DITU-net can offer a very close accuracy to ML based models, which further demonstrates the effectiveness of the proposed DITU-net based WPCM method.

Fig. 7 presents the WPCM results of the DITU-net and SNN considering two randomly selected WTs with different quantiles of the wind speed exclusion in IDP under S1, which means the model cannot access the data of the wind speed exceeding the set quantile. It is observable that the proposed DITU-net can produce more reasonable estimations of WPCs than SNN. Fig. 8 also shows the influence of the quantile settings on the performance of the proposed DITU-net and SNN under S2 for a fair comparison. It is observable that the proposed DITU-net can output stable WPCs in IDP; however, the performance of the SNN is diminished significantly in IDP.

Table V reports the average running time of each method. It is observable that the proposed DITU is faster than most of the considered benchmarks. Although the running time of the proposed DITU-net is 5 times longer than the SR, it is already acceptable for the online implementation. A regular PC with a Core i7-8700 CPU, a NVIDIA GeForce GTX 2080Ti GPU, and a 32Gb memory is utilized to conduct the computational experiments.

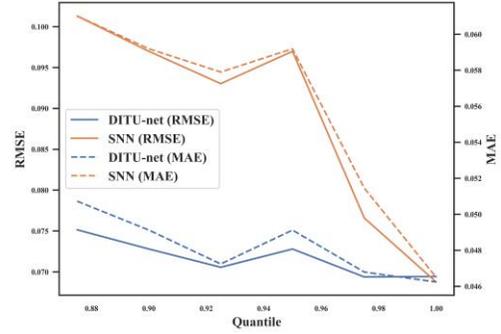

Fig. 8. Relationships between the quantile and the performance of DITU-net and SNN

*F. Discussions*

This research aims to develop one model offering a strong capability of being generalized to address WPCM for different WTs without retraining. In section IV.G, a large dataset with 76 WTs is employed to verify the effectiveness of the proposed DITU-net based method and results support that the proposed DITU-net could generalize well on considered WTs and achieve a high accuracy without data preprocessing and retraining. However, it is intractable to theoretically guarantee that the proposed DITU-net work well for all WTs in the world since training samples and considered deep generative networks also play a role in affecting its effectiveness. Thus, in the future, more datasets are needed to further verify the generalizability of the proposed DITU-net and inspire new developments based on the proposed DITU-net.

Meanwhile, compared with traditional WPCM methods, which typically need a repeated model fitting to derive WPCs from different SCADA datasets, our method shows a great potential of allowing one model to handle WPCM tasks of multiple wind turbines, which is already a significant advance.

## V. CONCLUSION

This paper presented a novel machine vision assisted WPCM method, the DITU-net. Firstly, we renovated the formulation of the WPCM problem from a machine vision perspective, based on which the DITU-net was developed to avoid sophisticated data preprocessing and retraining. The proposed DITU-net contained two steps, the DIT as well as the pixel mapping and correction. The DIT process was developed to approximate $f_{\Lambda_w}^{-1}$ with synthesized training image pairs, $I_x$ and $I_{wpc}$, via a U-net. After optimization, the developed U-net could be fed with the WPC images constructed based on observed SCADA data and offer the corresponding neat WPC $I_{wpc}$. To output a parametric form of the $I_{wpc}$, $f_{wpc}$, the pixel mapping and correction was developed. Firstly, based on the pixel mapping relationship, $I_{wpc}$ was converted to scattered SCADA data points and then fitted with a polynomial $f_p$. Next, the fitted $f_p$ was corrected based on the domain knowledge to ensure that the output $f_{wpc}$ could meet constraints of the WPC described in (1), (2.1) and (2.2).

Six classical benchmarks, three application scenarios and two application patterns were utilized to verify the effectiveness of the proposed DITU-net based on observed SCADA data collected from 76 WTs. The results demonstrated that, without any data preprocessing and retraining, the proposed DITU-net could offer the competitive accuracy with benchmarks applied with careful data preprocessing. The computational experiments also illustrated the superior of the proposed DITU-net for WPCM with insufficient SCADA data points.

In the future, we would like to improve the proposed DITU-net from two aspects. First, we would like to improve the DIT process by synthesizing more realistic data. Secondly, we would like to study a more advanced deep model architecture which better fit the WPCM problem.

## APPENDIX
### I. An illustration of synthesized SCADA WPC

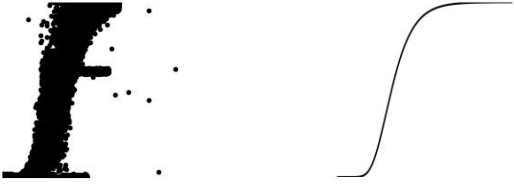

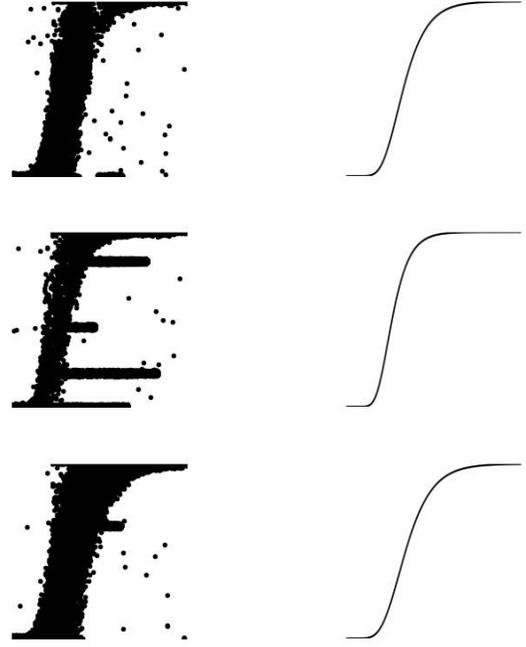

(a) SCADA WPC　　　　　(b) Neat WPC

Fig. A .1. An illustration of synthesized SCADA WPC and the corresponding synthesized neat WPC.

### II. Additional illustrations of WPCM results
More WPCM results are illustrated in Fig. A. 2.

### III. The analysis of the effectiveness of different distance-based loss functions in (11)

TABLE A. 1.
THE RESULTS OF DIFFERENT DISTANCE-BASED LOSS FUNCTIONS IN (11) BASED ON THE TESTING SET

| Evaluation Metrics | Loss function | | | Evaluation Metrics | Loss function | | |
|---|---|---|---|---|---|---|---|
| | MSE (11) | L1 Norm (30) | Smooth L1 Norm (31) | | MSE (11) | L1 Norm (30) | Smooth L1 Norm (31) |
| RMSE(S1) | 0.06962 | 0.07330 | 0.07140 | 0.05-SS (S1) (%) | 67.91 | 66.71 | 67.17 |
| RMSE(S2) | 0.06961 | 0.09231 | 0.07098 | 0.05-SS (S2) (%) | 67.85 | 66.63 | 67.34 |
| RMSE(S3) | 0.06893 | 0.08876 | 0.06924 | 0.05-SS (S3) (%) | 68.34 | 66.60 | 67.71 |
| MAE(S1) | 0.04688 | 0.04923 | 0.04803 | 0.10-SS (S1) (%) | 86.26 | 85.25 | 85.84 |
| MAE(S2) | 0.04684 | 0.06045 | 0.04797 | 0.10-SS (S2) (%) | 86.25 | 83.99 | 85.91 |
| MAE(S3) | 0.04616 | 0.05697 | 0.04694 | 0.10-SS (S3) (%) | 86.50 | 84.37 | 86.32 |
| MAPE(S1) (%) | 12.80 | 13.5 | 13.11 | 0.15-SS (S1) (%) | 93.91 | 93.14 | 93.53 |
| MAPE(S2) (%) | 12.82 | 15.95 | 13.13 | 0.15-SS (S2) (%) | 93.90 | 91.35 | 93.66 |
| MAPE(S3) (%) | 12.68 | 14.83 | 12.85 | 0.15-SS (S3) (%) | 94.09 | 91.87 | 94.04 |
| WMAPE (S1) (%) | 13.36 | 14.02 | 13.72 | | | | |
| WMAPE (S2) (%) | 13.35 | 16.68 | 13.69 | | | | |
| WMAPE (S3) (%) | 13.14 | 15.93 | 13.37 | | | | |

$$\mathcal{L} = \left\| I_{wpc} - Unet(I_x) \right\|_1 \tag{30}$$

$$\mathcal{L} = \begin{cases} \dfrac{1}{2}\left\| I_{wpc} - Unet(I_x) \right\|_2^2, & |I_{wpc} - Unet(I_x))| \leq 1 \\ \left\| I_{wpc} - Unet(I_x) \right\|_1 - 0.5, & \text{otherwise} \end{cases} \tag{31}$$

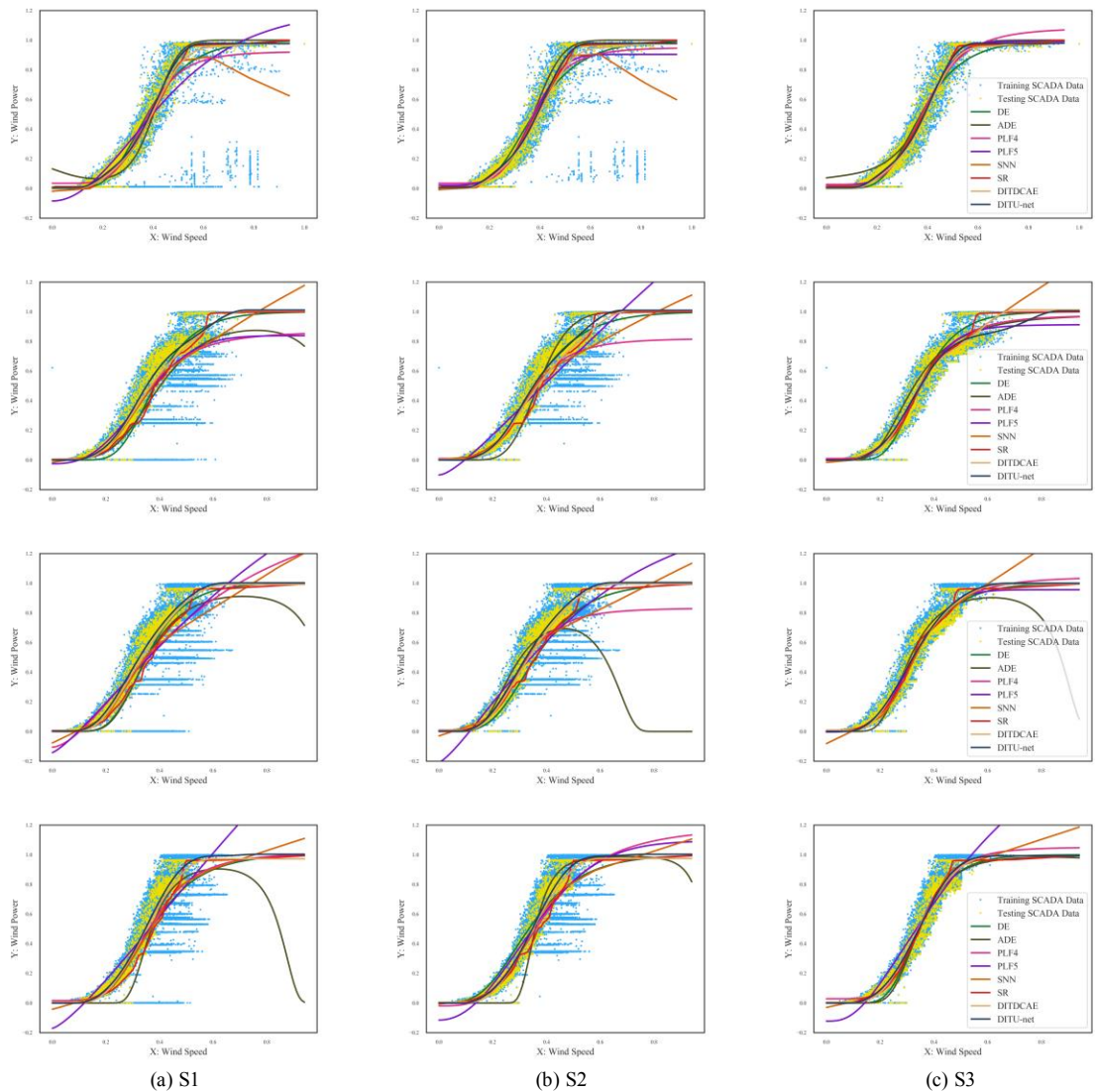

(a) S1                    (b) S2                    (c) S3

Fig. A. 2. Illustrations of WPCM results.

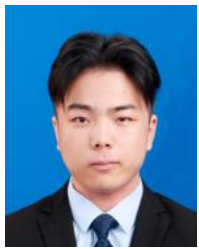

**Luoxiao Yang** received his B.E. degree in Electrical Engineering & Automation from Xi'an Jiaotong University, Xi'an, China, in 2019. He is current pursuing the Ph.D. degree in the School of Data Science at the City University of Hong Kong, Hong Kong, China. His research interests include data mining, renewable energy.

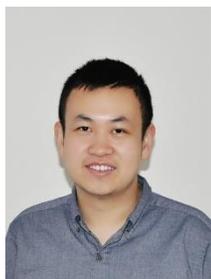

**Long Wang** (S'16-M'17) received the M.Sc. degree with distinction in computer science from University College London, London, U.K. in 2014, and the Ph.D. degree in systems engineering and engineering management from the City University of Hong Kong, Hong Kong, in 2017.

He is currently an Associate Professor with the Department of Computer Science and Technology, University of Science and Technology Beijing, Beijing, China. His research interests include machine learning, computational intelligence, computer vision, and their industrial applications. He was a recipient of the Hong Kong Ph.D. Fellowship, in 2014. He is a member of China Computer Federation (CCF), and CCF

Technical Committee on Computer Vision. He serves as an Associate Editor of IEEE Access and an Academic Editor of PLoS One, as well as a youth editorial committee member of Journal of Central South University. He is also a Lead Guest Editor of data science-related special issues on Measurement Science and Technology, Frontiers in Neurorobotics, Intelligent Automation & Soft Computing and Water.

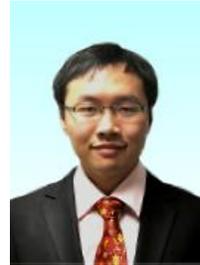

**Zijun Zhang** (M'12-SM'21) received his Ph.D. and M.S. degrees in Industrial Engineering from the University of Iowa, Iowa City, IA, USA, in 2012 and 2009, respectively, and B.Eng. degree in Systems Engineering and Engineering Management from the Chinese University of Hong Kong, Hong Kong SAR, China, in 2008.

Currently, he is an Associate Professor in the School of Data Science at City University of Hong Kong, Hong Kong SAR, China. His research focuses on machine learning and computational intelligence with applications in renewable energy, facility energy management, and intelligent transportation domains. He serves as an associate editor of IEEE Transactions on Sustainable Energy, IEEE Power Engineering Letters, and Journal of Intelligent Manufacturing.